\documentclass{book}
\usepackage{makeidx,epsfig}
\usepackage{setspace,graphicx}
\usepackage{Generic}
\usepackage{epsfig,epic,eepic,units}
\usepackage{hyperref}
\usepackage{url}
\usepackage{amsmath}
\usepackage{array}
\usepackage{algorithmic}
\usepackage{algorithm}
\usepackage{longtable}
\usepackage{mathrsfs}
\usepackage{mdwlist}
\usepackage{bbm}
\usepackage{multirow}
\usepackage{bigstrut}
\usepackage{amssymb}
\usepackage{graphicx}
\usepackage{bbm}
\usepackage[tight,footnotesize]{subfigure}
\makeindex

\begin{document}
\title{Automated OCT Segmentation for Images with DME}
\author{Sohini Roychowdhury, Dara D. Koozekanani, Michael Reinsbach and Keshab K. Parhi}

\pagenumbering{roman}
\begin{doublespace}
\maketitle 
\frontmatter
\tableofcontents
\mainmatter
\pagenumbering{arabic}

\section{Introduction}
Diabetic macular edema (DME) is a leading cause of vision loss in patients with diabetes. The World Health Organization estimates that by the year 2020, there will be 75 million blind people and 314 million partially blind people in the world \cite{WHO}. While treatments are available, including intra-vitreal injections and macular laser therapy, not all patients respond to these. Currently, there are no reliable methods for predicting patient response to therapy. Therefore, analysis of the patient images prior to treatment may allow the development of measures to predict patient response. Computer-aided diagnostic (CAD) systems enable automated detection of ophthalmic pathological sites, monitoring the progression of pathology and can guide follow-up treatment processes. Optical Coherence Tomography (OCT) images have been widely used to assess macular diseases, and they have enabled analysis of the extent of {\em disorganization} in the sub-retinal layers due to DME \cite{Sohini_cyst} \cite{Wilkins}. Sub-retinal layer disorganization refers to the variation in the underlying retinal micro-structure due to the presence of cystoid regions or due to disruptions in the cellular architecture of the sub-retinal layers due to pathology \cite{ref316}. For each patient's eye under analysis, a stack of images centered at the macula are acquired, such that reconstruction of the sub-retinal surfaces from the OCT image stacks aid localization of disease-related abnormalities in the retinal micro-structure. In this work, a CAD system is presented that automatically segments sub-retinal surfaces and layers in OCT image stacks from normal patients and abnormal ones, with DME, such that thickness maps corresponding to the sub-retinal layers can be further analyzed for their clinical relevance to the severity of DME.

Existing studies on OCT images from patients with DME by Maalej et al. \cite{ref2} have shown that clinical edema causes the tomographic thickness to increase by 1.33 times the normal thickness values. Also, significant relationship has been found between retinal thickness and visual acuity \cite{ref2}. Other studies by Otani et al. \cite{Otani} have shown strong correlation of visual acuity with the photoreceptor inner and outer segment junction integrity, and lower correlation between cystoid macular edema and visual acuity. Another study by Lattanzio et al. \cite{Lattanzio} shows that the macular thickness in eyes tend to increase with the severity of diabetic retinopathy and macular edema. In this work, we propose an automated CAD system that estimates the sub-retinal layer thicknesses in healthy and pathological OCT images with DME with correlation coefficient $r\geq0.7$. Such an automated system can lead to faster and more efficient detection and treatment of patients with DME if the macular thickness for the particular patient is found to exceed the clinically acceptable levels. Additionally, such an automated system can be used  for clinical trials to analyze more metrics other than the macular thickness for their clinical relevance to visual acuity.

The proposed automated OCT segmentation algorithm involves two key steps. In the first step, additive noise introduced by the imaging systems is removed from the images. In the second step, the denoised images are subjected to model-based segmentation. In prior efforts for denoising OCT images in the first step, OCT image noise suppression has been accomplished by weighted averaging in 3-D space by Mayer et al. \cite{OCTSEG}, discrete wavelet transforms by Adler et al. \cite{wavelet} and dual-tree wavelet transforms by Chitchian et al. \cite{Erik}. In another work by Fang et al. \cite{Farsiu_denoise}, multi-scale sparsity based tomography denoising is performed by sub-sampling the noisy OCT image, followed by dictionary parameter training, k-means clustering and final image restoration. It is observed that the sparsity-based denoising approach by Fang et al. \cite{Farsiu_denoise} is most useful in signal retrieval from noise in un-averaged OCT frames from Bioptigen SDOCT systems. However, images obtained from Spectralis OCT (Heidelberg Engineering, Germany) have built-in stabilization and averaging systems that produce significantly high SNR in the OCT images. In this work, we analyze images from Spectralis OCT and compare the denoising performances of the proposed Wiener deconvolution algorithm with Fourier-domain based noise variance estimator with the complex dual-tree wavelet transform method proposed by Chitchian et al. \cite{Erik}. This comparative analysis demonstrates the importance of image noise removal on automated OCT segmentation algorithms.

For the second step of automated sub-retinal surface and layer segmentation, some prior works rely on edge detection methods searching for peaks of gradient changes in column wise pixel intensities followed by interpolation by Koozekanani et al. \cite{Dara1}, or edge classification by Ishikawa et al. \cite{Ishikawa} and Bacgi et al. \cite{Bacgi}. Such edge detection methods, however, suffer from large segmentation errors when the image pixel intensities are inconsistent and in the presence of underlying pathology. Active contours based methods have been explored by Mishra et al. \cite{Mishra} with a two-step kernel based optimization scheme for sub-retinal layer segmentation. Another well known graph-cut segmentation method separates a 3-D composite image into sub-retinal layers by finding a minimum-cost closed set in a 3D graph by Garvin et al. \cite{Garvin}. Modifications to this method have been used to segment the optic nerve head by Antony et al. \cite{Bhavna}, and an alternative segmentation approach using dynamic programming have been explored by Chiu et al. \cite{Farsiu} and Lee at. al. \cite{Lee}. Further,  multi-resolution graph search has been applied for segmentation of up to 3 most significant sub-retinal surfaces for DME images by Abhishek et al. \cite{comp1}, and for segmenting up to 12 surfaces in OCT images with pigment epithelial detachment by Shi et al. \cite{comp2}. The major drawback of such graph-based 3-D segmentation methods is that they suffer from high computational complexity and are restrictive in their assumptions. Another well-known freely available OCT image segmentation system (OCTSEG) by Mayer et al. \cite{OCTSEG} utilizes second-order edge detection followed by fifth-order polynomial fitting for automated segmentation of the sub-retinal surfaces. Although this method results in fast OCT segmentation in normal images, the inner sub-retinal surfaces are incorrectly identified in abnormal OCT images with pathology. 

This work makes two key contributions. The first contribution is the comparative analysis of the existing wavelet-transform based denoising method with a novel Fourier-domain structural error-based denoising method presented in our previous work \cite{Sohini_asilomar}. In this previous conference paper, the concept of noise parameter estimation for Wiener deconvolution was described. In this work, we analyze the importance of image denoising on automated sub-retinal surface and layer segmentation process. Our analysis demonstrates that the wavelet-based denoising approach loses most sub-retinal surfaces by over smoothing the noisy OCT images, thereby incurring approximately two times more error in segmenting sub-retinal surfaces and layers when compared to the proposed Fourier-domain based denoising method. The second contribution is a novel multi-resolution iterative sub-retinal surface segmentation algorithm that is adaptive and robust to normal OCT images and images with DME. Also, the proposed method is comparatively analyzed with the OCTSEG system on normal and abnormal OCT images for automated sub-retinal surface segmentation performance. 

The organization of this chapter is as follows. In Section \ref{method}, the method and materials used to analyze the proposed automated segmentation system are presented. In Section \ref{result} the experimental results of automated sub-retinal surface and layer segmentation are presented. Conclusions and discussion are presented in Section \ref{conclusion}.

 \section{Materials and Method}\label{method}
\begin{figure}[ht]
\begin{center}
\includegraphics[width = 6.0in]{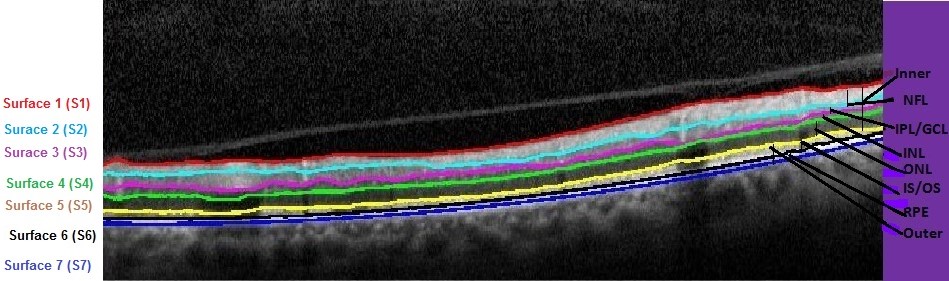}
\caption{The manually segmented Sub-retinal surfaces and layers in OCT images. The 7 sub-retinal surfaces are color coded as Surface 1 (Red), Surface 2 (Cyan), Surface 3 (Pink), Surface 4 (Green), Surface 5 (Yellow), Surface 6 (Black), Surface 7 (Blue). The sub-retinal layers are: NFL, IPL/GCL, INL, ONL, IS/OS, Inner and Outer layers.}\label{defination}
\end{center}
\end{figure}

The two key steps for automated denoising and segmentation of the sub-retinal layers from OCT images are described in the following subsections. From all the OCT data image stacks each OCT image is treated as a separate standalone image while denoising and segmenting the sub-retinal surfaces. This operation of separately segmenting each image without any information regarding the adjacent images from the OCT image stack is different from the existing graph-based segmentation approaches in \cite{Garvin}. Once each image is denoised, 7 sub-retinal surfaces, i.e., Surface 1 to Surface 7, and 6 sub-retinal layers that are then segmented are shown in Fig. \ref{defination}. The segmented sub-retinal layers extend from the inner limiting membrane (ILM, Surface 1) to the Bruch's Membrane (BM, Surface 7) surface. The automatically segmented sub-retinal layers as shown in Fig. \ref{defination} are: the Nerve Fiber Layer (NFL, between Surface 1 and Surface 2), Inner Plexiform Layer (IPL) and Ganglion Cell Layer (GCL) combined (IPL/GCL, between Surface 2 and Surface 3), Inner Nuclear Layer and outer plexiform layer combined (INL, between Surface 3 and Surface 4), Outer Nuclear Layer (ONL, between Surface 4 and Surface 5), Photoreceptor Inner/Outer Segment (IS/OS, between Surface 5 and Surface 6) and Retinal Pigment Epithelium (RPE, between Surface 6 and Surface 7). Additionally, to analyze the average retinal thickness two more combined layers are analyzed as: the Inner layer, that combines the NFL, IPL/GCL, INL and ONL (between Surface 1 and Surface 5), and the Outer layer, that combines the IS/OS and RPE layers (between Surface 5 and Surface 7). 

The data for assessing the performance of automated OCT segmentation is obtained using the Heidelberg Spectralis Imaging system at the Department of Ophthalmology of the University of Minnesota. OCT image stacks from normal and abnormal eyes with DME are obtained such that each image is averaged over 12-19 frames. Each OCT image has a resolution of 5.88 $\mu m/$pixel along the length and 3.87$\mu m/$pixel along the width and [1024x496] pixels per image. To evaluate the performance of the proposed segmentation system, each image is manually segmented by a human expert for all the sub-retinal surfaces. The following two sets of OCT image stacks are obtained corresponding to normal eyes and eyes with pathology.
\begin{enumerate}
\item Normal set: This image set comprises of OCT image stacks of the macular region from 10 healthy eyes with 19-25 images per OCT stack. A total of 203 normal OCT images are collected in this set.
\item Abnormal set: This image set comprises of macular OCT image stacks from 15 patients with DME and 19-33 images per OCT stack. A total of 357 images with sub-retinal cysts and layer disorganization are collected in this set. The Surface 6, which is not affected by DME, is not manually marked for images from this data set.
\end{enumerate}

\subsection{Automated Denoising}
Additive noise removal from OCT images is achieved based on the inherent assumption that the additive noise is Gaussian in nature. The statistical model for a noisy image ($I$) follows $\psi(I)=\psi(H).\psi(I_d)+\psi(N)$, where, $I$ is the noisy OCT image, $H$ is a 2-D point spread function (PSF), $I_d$ is the noiseless image, $N$ is the additive Gaussian noise and $\psi$ denotes a particular frequency-based or texture-based transform. Some existing OCT image denoising methods in \cite{Erik} \cite{wavelet} achieve image denoising using the above statistical model with the Wavelet-transform. In this work, we estimate additive noise using the Fourier-transform as function $\psi$ \cite{Sohini_asilomar}.

The proposed system utilizes the Wiener deconvolution algorithm that reduces the least-square error in Fourier-domain \cite{Wiener} as shown in (\ref{den}).
\begin{equation}\label{den}
\psi(I_{d})=\left[\frac{1}{\psi(H)}\frac{\left|\psi(H)\right|^2}{\left|\psi(H)\right|^2+ \frac{\Sigma_{N}}{\Sigma_{I_d}}}\right] \psi(I),
\end{equation}
where $\Sigma_N$ and $\Sigma_{I_d}$ represent the power spectral density of noise and the original image, respectively. Assuming a Gaussian PSF ($H$) with zero mean and variance $10^{-3}$ \cite{Wiener}, the best estimate of noise to signal ratio ($\frac{\Sigma_N}{\Sigma_{I_d}}$) will provide the best estimate of a denoised image. It is known that for OCT images the strength of additive noise varies with the number of averaged frames and across imaging systems. Thus, it is imperative to estimate the additive noise variance of each OCT image separately. 

\begin{figure}[ht]
\begin{center}
\subfigure[]{\includegraphics[width = 3.0in, height=2.0in]{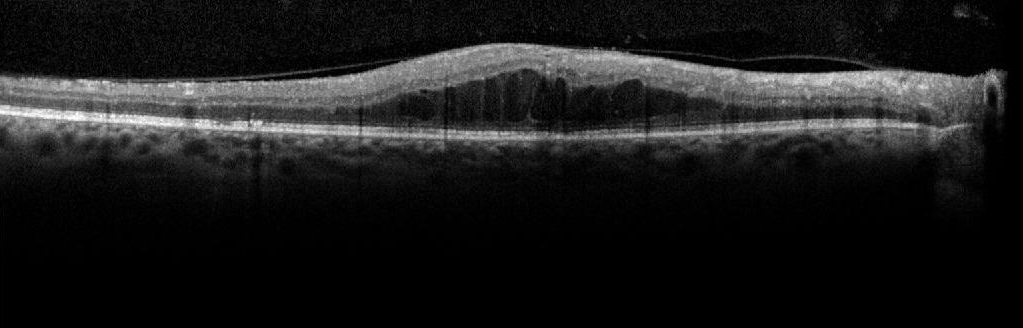}} 
\subfigure[]{\includegraphics[width = 3.0in,height=2.1in]{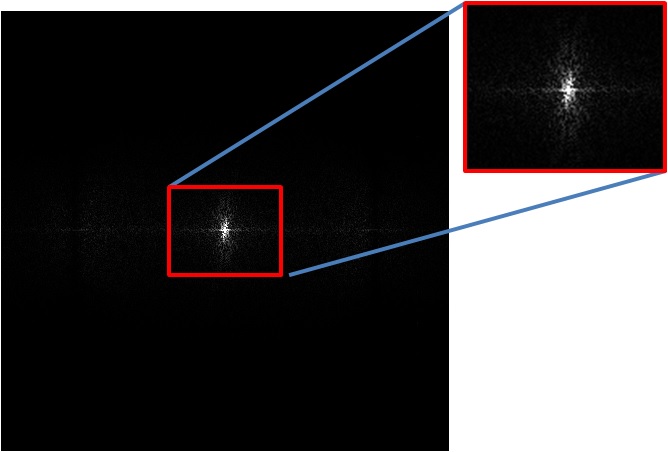}} 
\caption{Fourier-domain spectrum composition for OCT images. (a) Original noisy image ($I$) (b) Binary image of the Fourier-domain representation ($\psi'(I)$). The region within the red box is the central frequency spectrum structure that contains information regarding the sub-retinal micro-structure. $C$ is an image with components within in the red box only and all other pixels outside the box are zero. All remaining high-frequency components outside the red box are high-frequency terms introduced by noise and are represented by image $E$.}     \label{sample}
\end{center}
\end{figure}

In Fig. \ref{sample}, we observe that the absolute Fourier-domain representation of the noisy image ($|\psi(I)|$) has two prime components: the central frequency spectrum structure that contains information regarding the sub-retinal micro-structure ($C$), and the extra frequency components occupying high-frequency regions that appear due to the additive noise ($E$). Since the position of the frequency components in the Fourier-domain signify the sub-retinal components and the noise components, a binary image ($\psi'(I)$) can be generated from the Fourier domain representation of the noisy image $\psi(I)$, such that $\psi'(I)=|\psi(I)|>0$, and $\psi'(I)=C+E$. For an ideal denoised image, the central frequency spectrum must be conserved while the extra frequency terms introduced by noise must be suppressed. The challenge of separating the noise components from the central frequency components lies in the fact that the exact position of the sub-retinal and noisy components in the absolute Fourier-domain representation are unknown and variable across images.

\begin{figure}[ht]
\begin{center}
\includegraphics[width = 5.0in, height=5.1in]{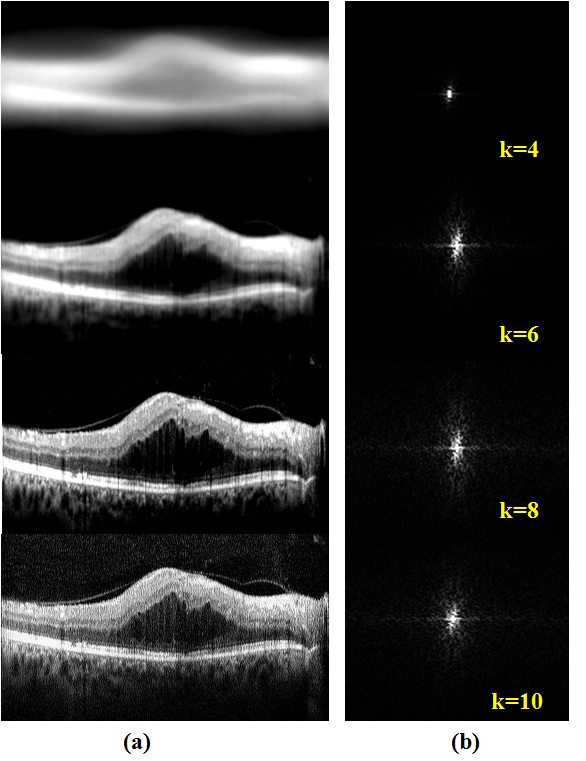}
\caption{Denoised OCT images by Wiener deconvolution with varying additive noise variance as $10^{-k}$. (a) Denoised image. (b) The respective absolute central frequency spectrum structure $C_k$ in the Fourier-domain representation of the denoised image when $k$ varies as $k=[4,6,8,9]$ and the additive Gaussian noise variance varies as $10^{-4}, 10^{-6}, 10^{-8}, 10^{-9}$, respectively. As estimated additive noise variance decreases from $10^{-4}$ to $10^{-9}$, the structure of $C_k$ becomes more similar to that of $C$, and additive noise components $E_k$ increases.}   \label{change}
\end{center}
\end{figure}

We observe that absolute Fourier-domain representations of the denoised images ($|\psi(I_{d_k})|$) can also be separated into the central frequency and the extra frequency components as $|\psi(I_{d_k})|=C_k+E_k$, where, $I_{d_k}$ is estimated by Wiener deconvolution using additive Gaussian noise estimate $N(0, 10^{-k})$. For an image with spatial configuration of [$n_1$x$n_2$] pixels, analysis of the Fourier-domain noise component image ($E_{k}$) demonstrates, $\sum_{i=1}^{n_1}\sum_{j=1}^{n_2}E_k(i,j)\approx 0$ for Wiener deconvolution with high-noise variances (around $10^{-1}$ to $10^{-8}$), and $\sum_{i=1}^{n_1}\sum_{j=1}^{n_2}E_k(i,j)>0$ for deconvolution with low-noise variances (around $10^{-10}$ to $10^{-15}$). Here, $(i,j)$ refers to the frequency domain image pixels. The Fourier-domain central frequency structure of the denoised images ($C_{k}$) with different noise variances is shown in Fig. \ref{change}. 

For the best estimate of additive noise variance, a Fourier-domain structural error metric ($e$) is defined in (\ref{eq1}-\ref{eq11}). A feasibility range for the convex error metric `$e$' can be identified by the range of values for a constant `$a$'.
\begin{eqnarray}\label{eq1}
e_k=\sum_{i=1}^{n_1}\sum_{j=1}^{n_2} \left[\psi'(I)-a(|\psi(I_{d_k})|.\psi'(I)) +\psi(I_{d_k})^2\right] (i,j)\\ \nonumber
\text{Since,~~}|\psi(I_{d_k})|=C_k+E_k, \psi'(I)=C+E,\\ 
e_k=\sum_{i=1}^{n_1}\sum_{j=1}^{n_2}[C+E-a(C.C_k+E.E_k)+C_k^2+E_k^2](i,j).\label{eq11}\\\nonumber
\end{eqnarray}

\begin{figure}[ht]
\begin{center}
\includegraphics[width = 4.0in, height=3in]{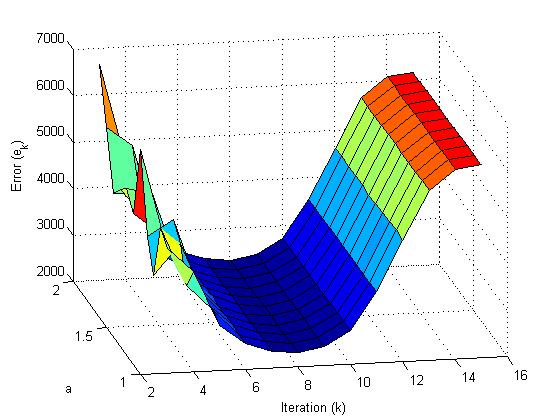}
\caption{Fourier-domain error $e_k$ as $k=[1,2...15]$ and $a=[1,1.1,1.2....2]$. Minimum error occurs at $a=1$ and $k=8$. Thus, best estimate of additive noise for this image is $N(0,10^{-8})$.}     \label{sample_error}
\end{center}
\end{figure}

Next, the trends of the structural error are analyzed in \eqref{eq22}-\eqref{eq22_6} by considering the situations when the estimated noise variance is high in \eqref{eq22_1}-\eqref{eq22_3} and low in \eqref{eq22_4}-\eqref{eq22_5}, respectively. From Fig. \ref{change}, it is evident that as the estimated Gaussian noise variance used for Wiener deconvolution iteratively decreases (as $10^{-k}$, $k={1,2,... 15}$), the central components $C_k$ of the denoised image become more similar to $C$, and the extra additive noise components increase. The iteration at which this Fourier-domain error is minimum corresponds to the best estimate of the additive Gaussian noise and Wiener deconvolution of noisy OCT image $I$ using this best noise variance estimate yields the best denoised image $I_d$. 
\begin{eqnarray}\label{eq22}
e_k-e_{k+1}=\sum_{i=1}^{n_1}\sum_{j=1}^{n_2}[a(C.C_{k+1}-C.C_{k}+E.E_{k+1}\\ \nonumber
-E.E_{k})+C_k^2-C_{k+1}^2+E_k^2-E_{k+1}^2](i,j).\\ \nonumber
{\bf High-noise~variance:}E_{k+1}\approx E_{k},\sum_{i=1}^{n_1}\sum_{j=1}^{n_2}E_{k}(i,j)\approx 0, \\ 
\text{Since~}C_k,E_k\in [0,1] \Rightarrow \forall (i,j), \text{and if~} a\geq1, \label{eq22_1}\\ \nonumber
a.C.C_k(i,j) \geq C^2_{k}(i,j),\sum_{i=1}^{n_1}\sum_{j=1}^{n_2}[a.C.C_{k+1}-C_{k+1}^2](i,j)\\ 
\geq \sum_{i=1}^{n_1}\sum_{j=1}^{n_2}[a.C.C_{k}-C_{k}^2](i,j).\label{eq22_2}\\
\Rightarrow e_k-e_{k+1}\geq 0 \Rightarrow e \text{ has decreasing trend}.\label{eq22_3}\\\nonumber
{\bf Low-noise~variance:~~~~~~~~~~} C_{k+1}\approx C_{k}, \text{and~~} \forall (i,j)\\ 
\sum_{i=1}^{n_1}\sum_{j=1}^{n_2} E.E_{k+1}(i,j) \approx \sum_{i=1}^{n_1}\sum_{j=1}^{n_2} E.E_k(i,j),\label{eq22_4}\\ \nonumber
\sum_{i=1}^{n_1}\sum_{j=1}^{n_2}E^2_{k+1}(i,j)\geq \sum_{i=1}^{n_1}\sum_{j=1}^{n_2}E^2_{k}(i,j)\geq 0, \\
\Rightarrow e_k-e_{k+1}\leq 0 \Rightarrow e \text{ has increasing trend.}\label{eq22_5}\\ \nonumber
\text{Substituting \eqref{eq22} and \eqref{eq22_4} in \eqref{eq22_5}~we~have:~}\forall (i,j),~~~~~~~\\ \nonumber
a([E.E_{k+1}-E.E_{k}](i,j))+[E^2_{k}-E^2_{k+1}](i,j) \leq 0\\ 
\Rightarrow a \leq [E_{k+1}+E_{k}](i,j)\Rightarrow a \leq 2.\label{eq22_6}\\ \nonumber
\end{eqnarray}
To ensure iterative optimality of $e_k$, it is necessary that $[1 \leq a \leq 2]$. Fig. \ref{sample_error} shows the Fourier-domain error $e$ as iteration $k$ and constant $a$ varies. This particular image has low variability with respect to $a$. For all OCT images denoised using the proposed approach, the error $e_k$ is minimized with respect to $a$ and $k$ to obtain the best denoised image $I_d$.

To comparatively assess the importance of noise removal on automated OCT image segmentation, we selected a baseline method of image denoising by the wavelet tree-based approach in \cite{Erik} \cite{wavelet_ext}. Wavelet-based denoising approaches are known to estimate clean signal coefficients from the noisy ones using Bayesian estimators \cite{wavelet_ext}. While wavelet coefficients with large magnitudes represent edges or textures, coefficients with small magnitude represent the smooth background regions. Thus, OCT image denoising using the wavelet shrinkage method by Chitchian et al. \cite{Erik} is achieved in four steps that include:  wavelet transform on the noisy image $I$, estimation of adaptive threshold using the wavelet transform, denoising image $I$ using the estimated threshold, and finally inverse wavelet transform on the denoised image. The most critical step in this wavelet-based denoising approach is the adaptive threshold estimation. While a low threshold will not produce significant denoising, a high threshold will estimate a signal with a large number of zero coefficients that will correspond to over smoothened image with no edges or textures. To facilitate optimal threshold estimation, soft-thresholding is performed as described by Wagner et al. \cite{wavelet_ext}.  This method of OCT image denoising using complex dual-tree wavelet transform is referred as `CDWT' in the following sections. The performances of OCT image segmentation and sub-retinal layer thickness map estimation using the proposed denoising approach and the wavelet-based denoising approach are compared. The pixel intensities of the denoised image ($I_d$) are scaled in the range [0,1] before automated segmentation.

\subsection{Automated Segmentation}
Every denoised image ($I_d$) is subjected to multi-resolution iterative high-pass filtering to extract 7 sub-retinal surfaces \cite{Sohini_asilomar}. Several OCT image segmentation methods that involve edge detection followed by segment linking or gradient search methods have been developed \cite{Dara1}. However, most such methods fail in the presence of sub-retinal disorganization owing to pathology. Since the proposed method of sub-retinal surface segmentation does not require interpolation between partially detected line segments, it is robust to normal and abnormal images with pathology.

Each sub-retinal layer in the OCT images has a distinct pixel intensity that is different from the adjacent layers. Also, some sub-retinal surfaces are more easily detectable than the others; for instance the top surface of the NFL (Surface 1) and the IS/OS junction (Surface 5) are easily detectable when compared to the other surfaces. Thus, to capture each sub-retinal surface separately, it is imperative to look for the most distinct edges/surfaces first, then mask them out, and analyze the remaining regions for the next most significant surface till all 7 surfaces have been detected. This idea is the basis for the proposed multi-resolution image segmentation approach, where the sub-retinal surfaces are segmented iteratively in decreasing order of their detectability, with the most detectable surface (Surface 1) segmented first and the least detectable surface (Surface 2) segmented at last.

The proposed multi-resolution surface extraction algorithm proceeds in 6 iterative steps, i.e., $l=[1,2,...6]$. In each iterative step `$l$' a different source image is used. The source image ($I_{s_l}$) refers either to the denoised image or the negative of the denoised image, whichever enhances the detectability of the surface of interest. Next, a region of interest is selected for the source image based on the region where the most distinctive edge will lie. This region of interest is denoted by a masked region $G_l$. Next, the region in source image $I_{s_{l}}$ that lies within masked region $G_l$ is high-pass filtered in different ways as shown in (\ref{eqn3}). The `$\circ$' operator denotes pixel-wise multiplication. The filtered image ($I_k$) is then subjected to pixel-level thresholding ($\Delta_k$) and the regions in the remaining image are subjected to a particular decision criterion ($\phi_l$) to extract a binary image ($I_{r_l}$) with one or more regions in it that satisfy the segmentation decision criterion in \eqref{eq3_1}. The top or bottom surface ({\em Surf}$_l$) of the region/regions in image $I_{r_l}$ is determined as the segmented surface ($S_l$) in \eqref{eq3_2}. 
\begin{eqnarray}\label{eqn3}
\forall l=[1,2...6], I_{l}=HPF_{l}(I_{s_l} \circ G_l)\\ \nonumber
\Rightarrow I_{l}=I_{s_l} \circ G_l -LPF_l(I_{s_l} \circ G_l).\\
I_{r_l}=\phi_l(\Delta_l(I_l)).\label{eq3_1}\\
S_l=Surf_l(I_{r_l}).\label{eq3_2}\\ \nonumber
\end{eqnarray}
Corresponding to each iterative step ($l$), the choice of source image ($I_{s_{l}}$), mask for the region of interest ($G_l$), filter ($HPF_l/LPF_l$), threshold ($\Delta_l$), decision criterion ($\phi_l$) and surface of interest ({\em Surf}$_l$) are given in Table \ref{seg_it}. The decision criterion ($\phi_l$) for selecting a particular region of interest from the thresholded binary image $\Delta_l(I_l)$ can vary from the region with maximum area to the region with maximum major axis length as shown in Table \ref{seg_it}.

\begin{table*}[ht]
\begin{center}
\caption{Iterative parameters for sub-retinal surface segmentation using the proposed multi-resolution high-pass filtering method. These parameters are used in \eqref{eqn3}-\eqref{eq3_2} to obtain 7 sub-retinal surfaces in 6 iterations.}
\resizebox{\textwidth}{!}{\begin{tabular}{|c |c |c |c|c|c|c|}
\hline 
Iteration ($l$)&Source ($I_{s_{l}}$)&Mask ($G_l$)&Filter ($HPF_l/LPF_l$)&Threshold ($\Delta_l$)&Criterion ($\phi_l$)&{\em Surf}$_l$\\ \hline
1&$I_d$&whole image&$HPF_1=$2-D Gaussians,&regiongrow (${I_1}$ in [0, 255],&$\arg\max$ (Area)&$S_1=$ Top ($I_{r_1}$), \\
&&&horizontal and vertical& seed=255, threshold$=\left\{230-235\right\})$&&choroid$=$ Bottom ($I_{r_1}$).\\ \hline
2&$1-I_d$&Between $S_1$&$LPF_2=$average [25x25]&pixel values $<0$&$\arg\max$ (Major&$S_5=$Top ($I_{r_2}$),\\
&&and choroid&&& Axis Length)&$S_7=$Bottom ($I_{r_2}$).\\ \hline
3&$1-I_d$&Between $S_5$&$HPF_3=$contrast &pixel values $>$Otsu's Threshold&-&$S_6=$Bottom ($I_{r_3}$).\\
&&and $S_7$&enhancement&estimated on $I_3$&&\\ \hline
4&$1-I_d$&Between $S_1$&$LPF_4=$average [25x25]&pixel values $>$Otsu's Threshold&$\arg\max$ (Major&$S_4=$Bottom ($I_{r_4}$),\\
&&and $S_5$&&estimated on $I_4$&Axis Length)&.\\ \hline
5&$1-I_d$&Between $S_1$&$LPF_5=$2-D Gaussian,&pixel values $>$Otsu's Threshold&$\arg\max$ (Area)&$S_3=$Bottom ($I_{r_5}$),\\
&&and $S_4$&horizontal [10x10]&estimated on $I_5$&&\\ \hline
6&$I_d$&Between $S_1$&$LPF_6=$average [25x25]&pixel values $>$Otsu's Threshold&$\arg\max$ (Area)&$S_2=$Bottom ($I_{r_6}$),\\
&&and $S_3$&&estimated on $I_6$&&\\ \hline
\end{tabular}}
	\label{seg_it}
	\end{center}
\end{table*} 

\begin{figure}[ht]
\begin{center}
\subfigure[]{\includegraphics[width = 3.1in,height=1.7in]{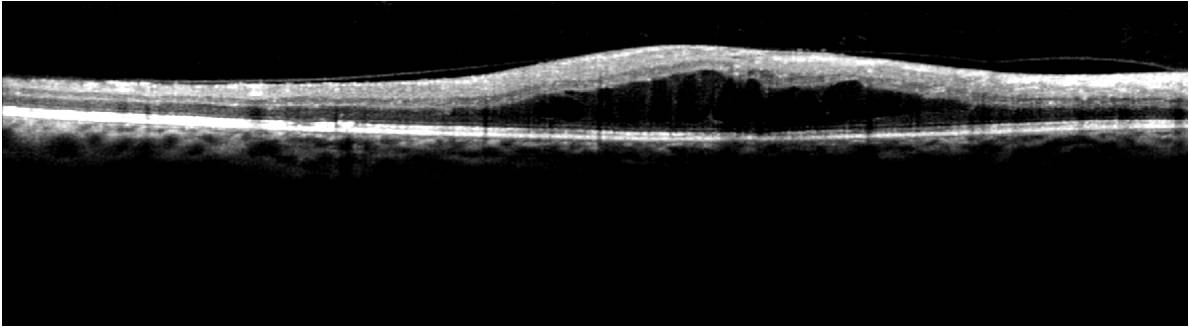}}
\subfigure[]{\includegraphics[width = 3.1in,height=1.7in]{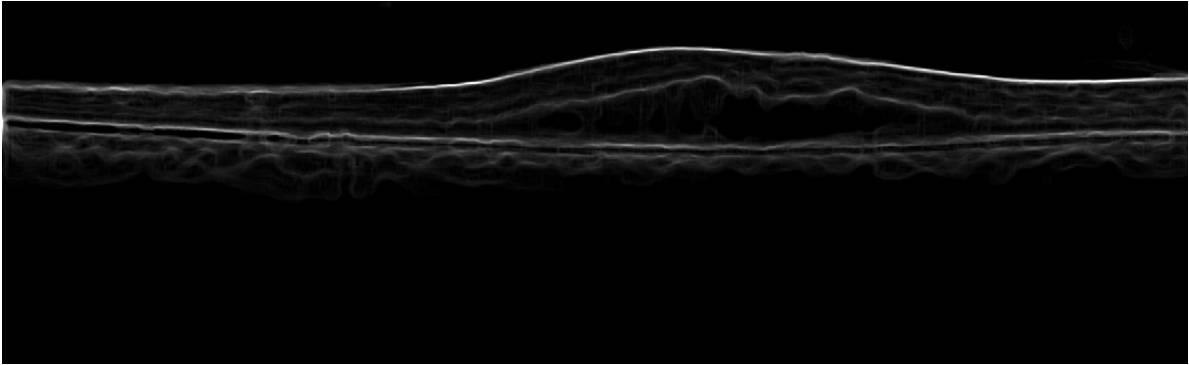}}
\subfigure[]{\includegraphics[width = 3.1in,height=1.7in]{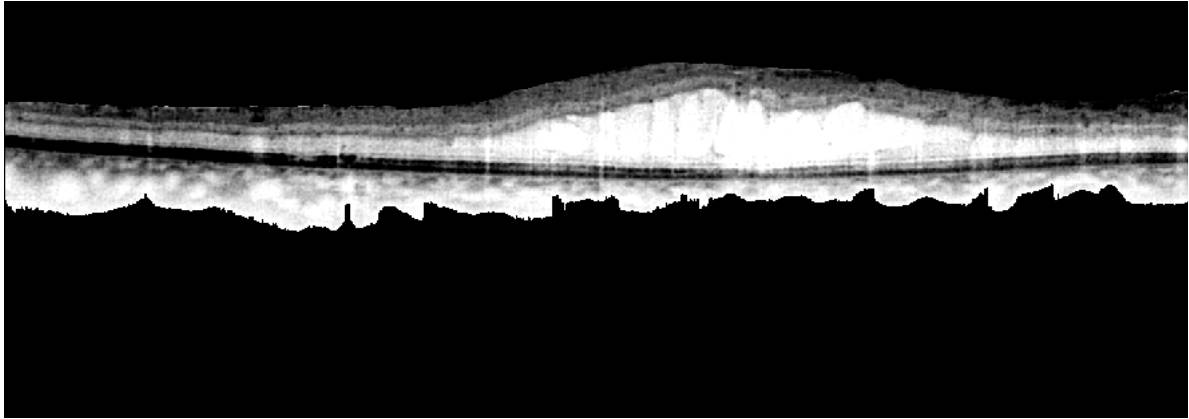}}
\subfigure[]{\includegraphics[width = 3.1in,height=1.7in]{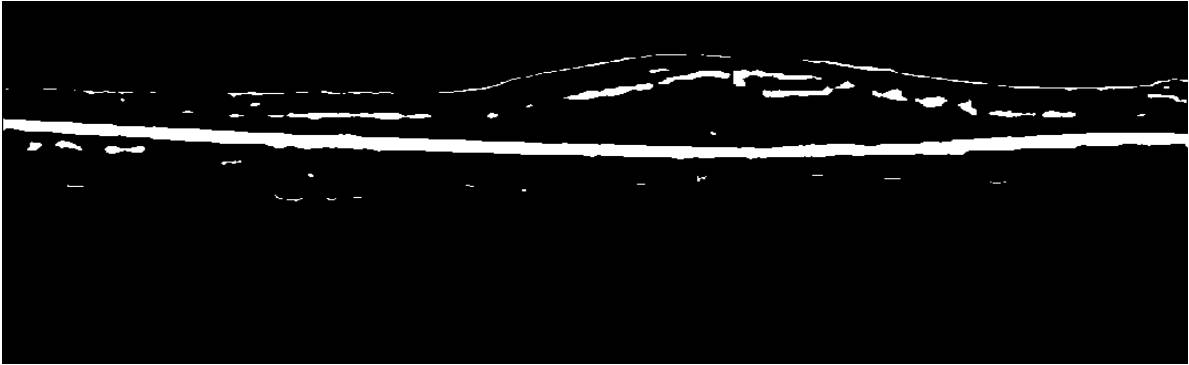}}
\subfigure[]{\includegraphics[width = 3.1in,height=1.7in]{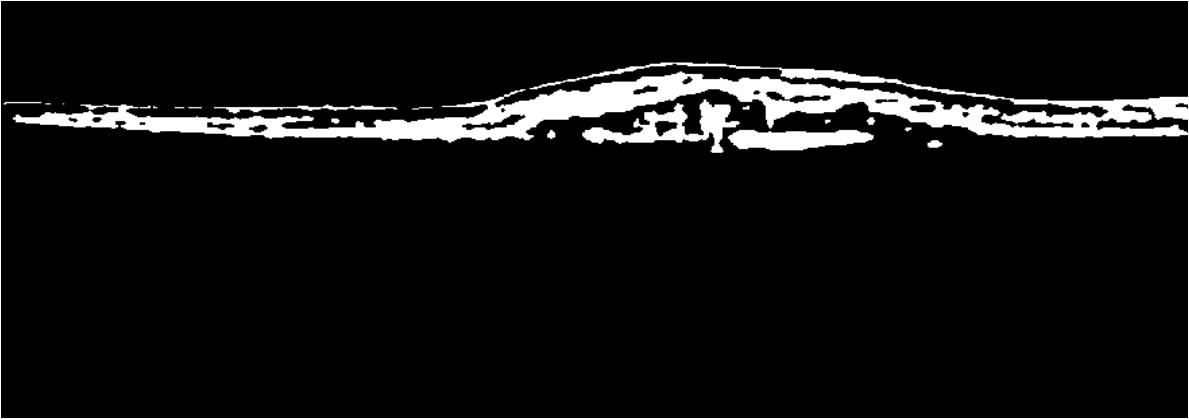}}
\subfigure[]{\includegraphics[width = 3.1in,height=1.7in]{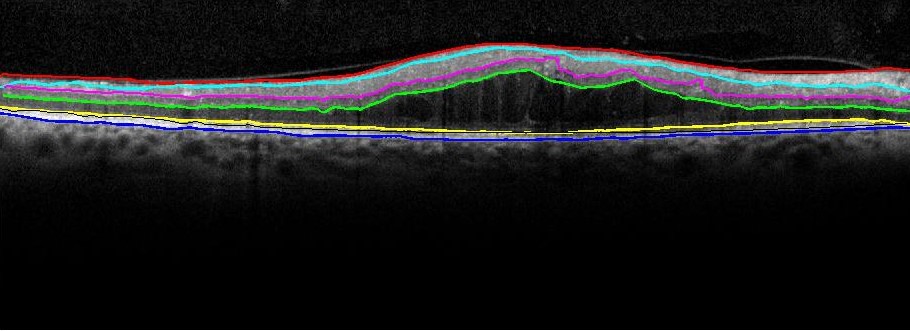}}
\caption{Examples of results at each step of the proposed iterative multi-resolution segmentation algorithm on an abnormal OCT image.}     \label{oct_seg}
\end{center}
\end{figure}

For an image from the Abnormal data set OCT stack, the proposed iterative surface segmentation algorithm is shown in Fig. \ref{oct_seg}. The denoised image obtained by the proposed approach ($I_d$) is shown in Fig. \ref{oct_seg}(a). The image $I_1$ obtained after high-pass filtering in iteration $l=1$ is in Fig. \ref{oct_seg}(b). Thresholding $I_1$ results in the detection of Surface 1 and the choroidal segment. In Fig. \ref{oct_seg}(c), negative source image $1-I_d$ in iteration $l=2$ is shown within the region of interest marked by $G_2$ that extends between the Surface 1 and the choroid segment. In Fig. \ref{oct_seg}(d), the image obtained after high-pass filtering and thresholding the image in Fig. \ref{oct_seg}(c) is shown. The region with maximum major axis length is extracted into image $I_{r_2}$. The top surface of this region is Surface 5, and the bottom surface is Surface 7. In Fig. \ref{oct_seg}(e), the image obtained in iteration $l=4$ after high-pass filtering and thresholding is shown. The region with maximum major axis length is selected in image $I_{r_4}$, and the bottom surface of this region is Surface 4. Two more iterations are performed to extract all 7 surfaces. In Fig. \ref{oct_seg}(f), automated segmentations achieved at the end of 6 iteration steps by the proposed method are shown. 

We observe that due to the multi-resolution nature of the proposed segmentation procedure, the inner sub-retinal surfaces ($S_2$ to $S_4$) are correctly detected in spite of the sub-retinal disorganization that occurs due to the presence of large cystoid regions. It is noteworthy that in certain OCT images, like the ones with the fovea as shown in Fig. \ref{defination}, while the Surface 1 is always continuous, other surfaces, such as Surface 3 or Surface 2, may appear discontinuous. To enable the detection of complete surfaces in such cases, an exception rule is applied to detect all surfaces, apart from Surface 1, that have chances of appearing discontinuous. For this exception rule, the Surface 1 from iteration $l=1$ is first detected and all its x-coordinates are noted. Next, from iteration 2 through 6, for each extracted surface, if the length of the surface is less than 75\% of the length of Surface 1, then another region satisfying the criterion ($\phi_l$) apart from the region already selected is added to image $I_{r_l}$. The final detected surface becomes the combination of surfaces from all the regions thus detected in image $I_{r_l}$ .

\section{Experiments and Results}\label{result}
The performance characteristics of automated segmentation of sub-retinal surfaces and sub-retinal layers are evaluated in three sets of experiments. In the first experiment, the performance of image denoising is analyzed for normal and abnormal OCT image stacks. In the second experiment, the error in segmenting the sub-retinal surfaces is analyzed for normal and abnormal images. In the third experiment, the mean sub-retinal layer thickness estimates obtained using automated algorithms are compared against the manual segmentation estimates. The impact of denoising on automated segmentation is evaluated by comparing the segmentation performance of the proposed Fourier-domain based denoising approach with the CDWT approach followed by the proposed multi-resolution iterative segmentation algorithm for sub-retinal surface and layer segmentation. 

\subsection{Performance of Automated Denoising}
For this experiment, the denoised OCT images are scaled in [0, 255] and then analyzed for pixel intensities. The performance metrics for evaluating the improvement in image quality by automated denoising are defined in terms of the image foreground region and background region. 
For an OCT image, the region of interest is the retinal micro-structure that lies between the ILM (Surface 1) and BM (Surface 7). The foreground region extends from 10 pixels above the manually segmented Surface 1 through 50 pixels below manually segmented Surface 7 to include the retinal micro-structure and the choroidal region. The remaining regions in each image are selected as the background. Next, $\mu_f, \sigma_f$  and $ \mu_b, \sigma_b$ are estimated as mean and standard deviation of pixel values in the foreground and background regions, respectively. The spatial denoising metrics are defined as global signal to noise ratio (SNR) in (\ref{eqf}), global contrast to noise ratio (CNR) in \eqref{eqf_1} and peak SNR (PSNR) in \eqref{eqf_2}. 

\begin{eqnarray}\label{eqf}
&SNR=20\log_{10}\frac{\mu_f}{\sigma_b}.\\ 
&CNR=\frac{\left|\mu_f-\mu_{b} \right|}{\sqrt{0.5\left(\sigma_f^2+\sigma_b^2\right)}}.\label{eqf_1}\\ 
&\forall [i',j'] \in I,I_d,\label{eqf_2}\\ \nonumber
&PSNR=20\log_{10}\left(\frac{\max_{i',j'}I(i',j')}{\sqrt {\frac{1}{n_{1}.n_{2}} \sum_{i'=1}^{n_{1}}\sum_{j'=1}^{n_2}\left(I(i',j')-I_d(i',j')\right)^2}}\right).\\ \nonumber
\end{eqnarray}

\begin{figure}[ht]
\begin{center}
\subfigure[]{\includegraphics[width = 3.0in,height=1.75in]{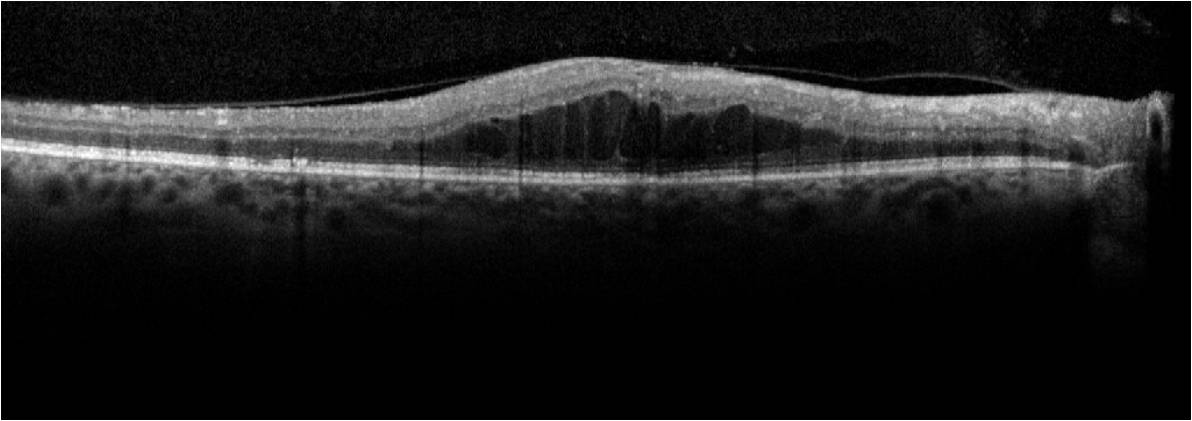}}
\subfigure[]{\includegraphics[width = 3.0in,height=1.75in]{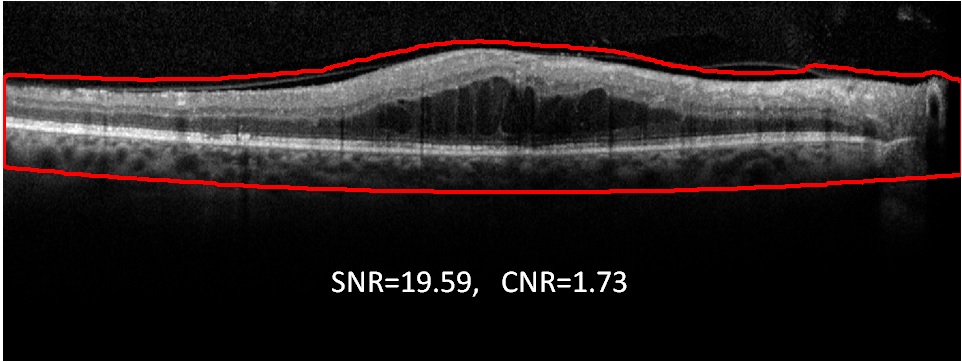}}
\subfigure[]{\includegraphics[width = 3.0in,height=1.75in]{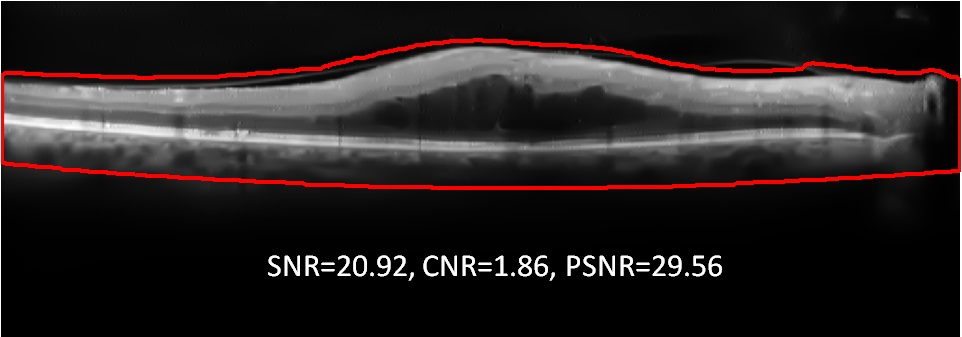}}
\subfigure[]{\includegraphics[width = 3.0in,height=1.75in]{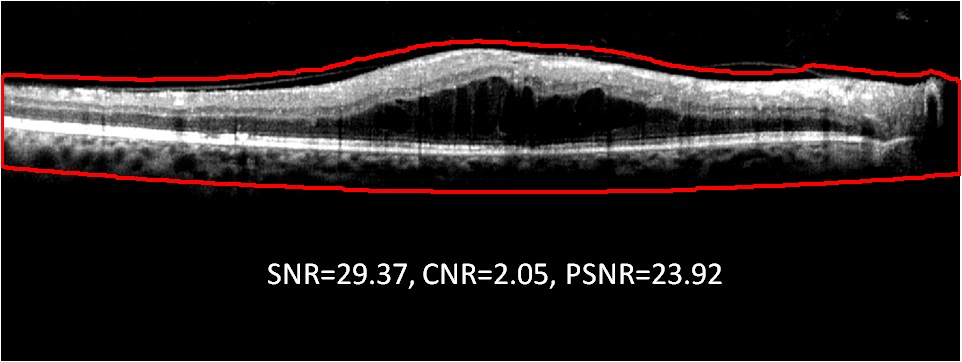}}
\caption{Example of automated OCT image denoising. (a) Noisy image ($I$), (b) The foreground region lies within the region bordered by the red boundary. All other regions are the background. (c) Denoised image by CDWT method. (d) Denoised image by the proposed method.}     \label{denoise_im}
\end{center}
\end{figure}

The foreground and background regions for an abnormal OCT image and the denoising performance metrics using the proposed Fourier-domain error based method and the wavelet-based CDWT approach are shown in Fig. \ref{denoise_im}. Here, we observe that the proposed denoising method retains the surface edges and contrast variation in the foreground region while it significantly reduces image noise from the background region when compared to the denoised image using CDWT. 

The comparison between denoising metrics using the proposed method and the CDWT method are shown in Table \ref{denoise_res}. Here, we observe that for OCT images from the normal and abnormal data sets, the proposed denoising method improves the image SNR by 12-13 dB, while the CDWT method improves the SNR by 3-4 dB only. The improvements in CNR fraction by the proposed and CDWT method are about 0.5 and 0.1, respectively. However, the CDWT method achieves PSNR about 4.5 dB greater than the proposed method. The PSNR of the proposed method is smaller than the CDWT method because of the denominator in (1.16) that denotes the similarity in structure between the denoised and noisy image, and a more similar structure will ensure a small value for this quantity. The proposed method alters the structure of the noisy image in the background region due to the underlying Fourier-domain error based Wiener deconvolution strategy, while the CDWT method conserves the image structure, and hence the difference in PSNR. The impact of the gain in SNR by the proposed denoising method versus the gain in PSNR by the CDWT method on automated sub-retinal surface segmentations are analyzed in the next experiments.
\begin{table}[ht]
\begin{center}
\caption{Performance of OCT image denoising using the proposed method versus the CDWT denoising method evaluated on normal and abnormal OCT image stacks.}
\begin{tabular}{|c |c |c |c|}
\hline 
Image&SNR (dB)&CNR&PSNR (dB)\\ \hline \hline
Normal set&& &\\ \hline
Noisy image&17.52$\pm$1.77&2.11$\pm$0.51&-\\ \hline
CDWT Denoising&21.03$\pm$1.87&2.20$\pm$0.54&27.91$\pm$1.39\\ \hline
Proposed Denoising&30.17$\pm$5.19&2.58$\pm$0.53&23.56$\pm$0.51\\ \hline \hline
Abnormal set&& &\\ \hline
Noisy image&16.92$\pm$1.74&1.79$\pm$0.105&-\\ \hline
CDWT Denoising&19.32$\pm$1.16&1.91$\pm$0.08&28.08$\pm$1.84\\ \hline
Proposed Denoising&29.23$\pm$2.23&2.32$\pm$0.11&23.67$\pm$0.38\\ \hline
\end{tabular}
	\label{denoise_res}
	\end{center}
\end{table}

\subsection{Sub-retinal Surface Segmentation Error}
The absolute errors incurred in segmenting each sub-retinal surface using the proposed denoising method or CDWT method followed by the proposed segmentation method on the normal and abnormal OCT images are shown in Table \ref{error_norm} and Table \ref{error_dme}, respectively. The mean and standard deviation in the absolute errors between the automatically segmented surface and the manually segmented surfaces for all OCT images under analysis are evaluated in $\mu m$. In Table \ref{error_norm}, the segmentation errors incurred by three existing segmentation algorithms on the normal images are also shown for comparison. The absolute surface segmentation error was estimated by Bagci et al. \cite{Bacgi} on 14 SD-OCT images, by Chiu et al. \cite{Farsiu} on 100 B-scans, and by Yang et al. \cite{Yang} on 38 image scans, respectively. 
\begin{table}[ht]
\begin{center}
\caption{Mean and standard deviation of sub-retinal surface segmentation error using the proposed method, and the CDWT method compared to the performance of existing methods on normal OCT images. These errors are computed in $\mu m$.}
\begin{tabular}{|c |c |c |c|c|c|c|}
\hline 
Surface&Proposed&CDWT&Bagci et al. \cite{Bacgi}&Chiu et al. \cite{Farsiu}&Yang et al. \cite{Yang}\\ \hline \hline
S1&1.22$\pm$0.96&8.45$\pm$3.92&4.3$\pm$0.8&2.8$\pm$2.3&2.2$\pm$0.7\\ \hline
S2&5.80$\pm$4.47&14.77$\pm$4.34&5.7$\pm$0.7&2.5$\pm$2.0&4.3$\pm$1.0\\ \hline
S3&2.78$\pm$1.53&13.38$\pm$6.14&5.3$\pm$0.5&3.2$\pm$2.4&3.2$\pm$0.9\\ \hline
S4&4.55$\pm$2.15&9.55$\pm$3.89&6.1$\pm$1.0&4.8$\pm$3.4&-\\ \hline
S5&1.71$\pm$1.38&4.27$\pm$2.17&8.8$\pm$1.2&3.8$\pm$2.9&-\\ \hline
S6&2.51$\pm$2.35&12.90$\pm$3.13&4.3$\pm$1.1&2.8$\pm$2.4&-\\ \hline
S7&2.05$\pm$2.17&12.58$\pm$5.01&5.5$\pm$1.0&3.2$\pm$2.8&2.4$\pm$1.2\\ \hline
\end{tabular}
	\label{error_norm}
	\end{center}
\end{table}

In Table \ref{error_norm}, we observe that on normal OCT images, the proposed denoising and segmentation method incurs 1-6 $\mu m$ of error across all the sub-retinal surfaces. This is significantly better than the CDWT denoising method followed by proposed segmentation that results in 4-14 $\mu m$ of error. Also, the proposed method has lower segmentation error when compared to other segmentation algorithms on all surfaces other than Surface 2, which is the least distinctive edge. Thus, on normal OCT images, the proposed method has relatively better sub-retinal surface segmentation performance when compared to CDWT or existing algorithms.
\begin{table}[ht]
\begin{center}
\caption{Mean and standard deviation of sub-retinal surface segmentation error using the proposed method, and the CDWT denoising method on abnormal OCT images. These errors are computed in $\mu m$.}
\begin{tabular}{|c |c |c |}
\hline 
Surface&Proposed&CDWT\\ \hline \hline
S1&4.02$\pm$4.82&10.97$\pm$5.64\\ \hline
S2&26.99$\pm$10.04&22.57$\pm$5.69\\ \hline
S3&19.97$\pm$7.49&23.64$\pm$12.18\\ \hline
S4&16.83$\pm$8.77&20.95$\pm$11.59\\ \hline
S5&6.81$\pm$6.42&8.02$\pm$2.29\\ \hline
S7&3.51$\pm$2.86&7.05$\pm$1.78\\ \hline
\end{tabular}
	\label{error_dme}
	\end{center}
\end{table}

For images from the abnormal set, the surface segmentation errors by the proposed method and CDWT method are shown in Table \ref{error_dme}. Since pathological changes due to DME do not affect the IS/OS layer (Surface 6), this sub-retinal surface was not manually segmented. For all the other sub-retinal surfaces, we observe a significant increase in the segmentation error for Surface 2, Surface 3, Surface 4 and Surface 5 for the abnormal OCT images when compared to the normal images. This increase in segmentation error is unavoidable since the disorganization caused by pathology significantly alters the inner retinal layers, making the inner surfaces significantly less distinct even for manual segmentation. In Table \ref{error_dme}, we also observe that for all surfaces other than Surface 2, the CDWT method incurs more error on abnormal images than the proposed approach. This happens because for abnormal images with large cysts, the Surface 2 may appear very close to Surface 1, in which case false edges may get detected by the proposed denoising method, causing the segmentation error to increase. In such cases the CDWT approach smoothens the false edges close to the Surface 1, thereby resulting in less error for Surface 2, but in turn increasing the error for segmenting Surface 3 through Surface 7 due to this smoothing action.

\begin{figure}[ht]
\begin{center}
\subfigure[]{\includegraphics[width = 3.0in,height=1.0in]{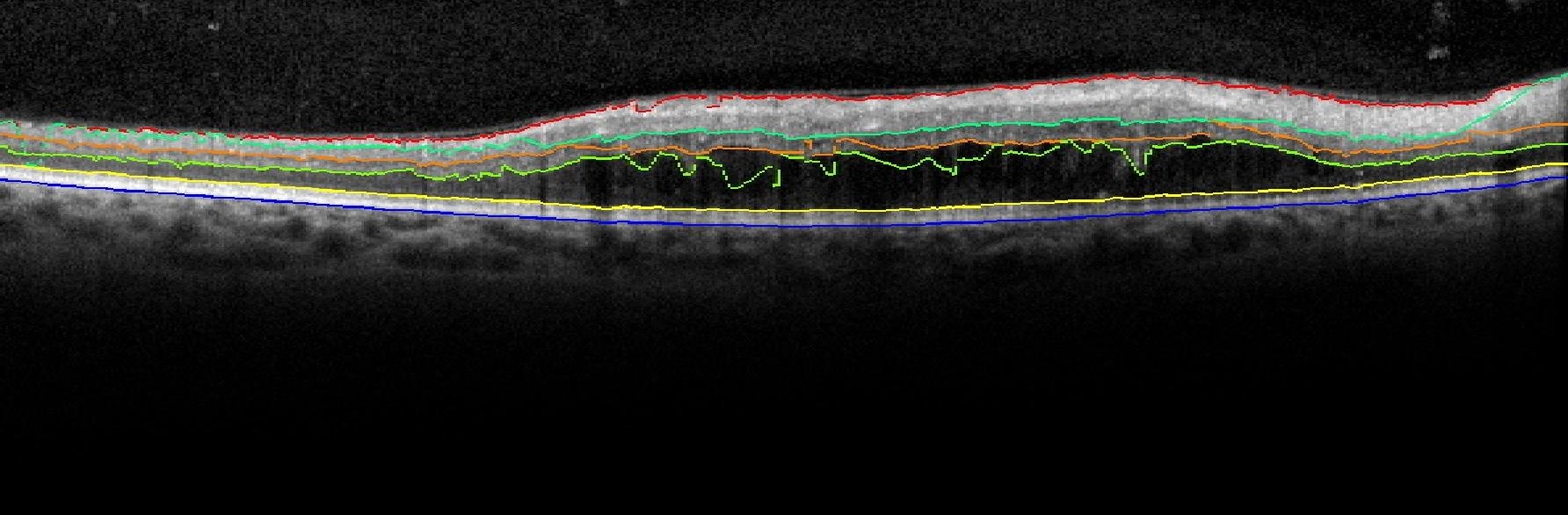}}
\subfigure[]{\includegraphics[width = 3.0in,height=1.0in]{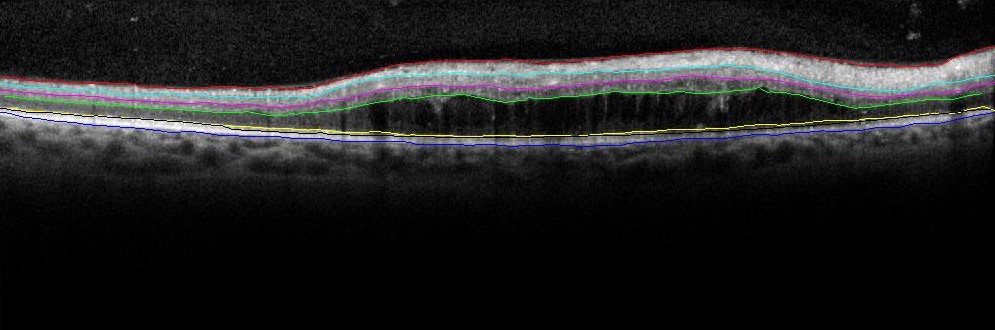}}
\subfigure[]{\includegraphics[width = 3.0in,height=1.0in]{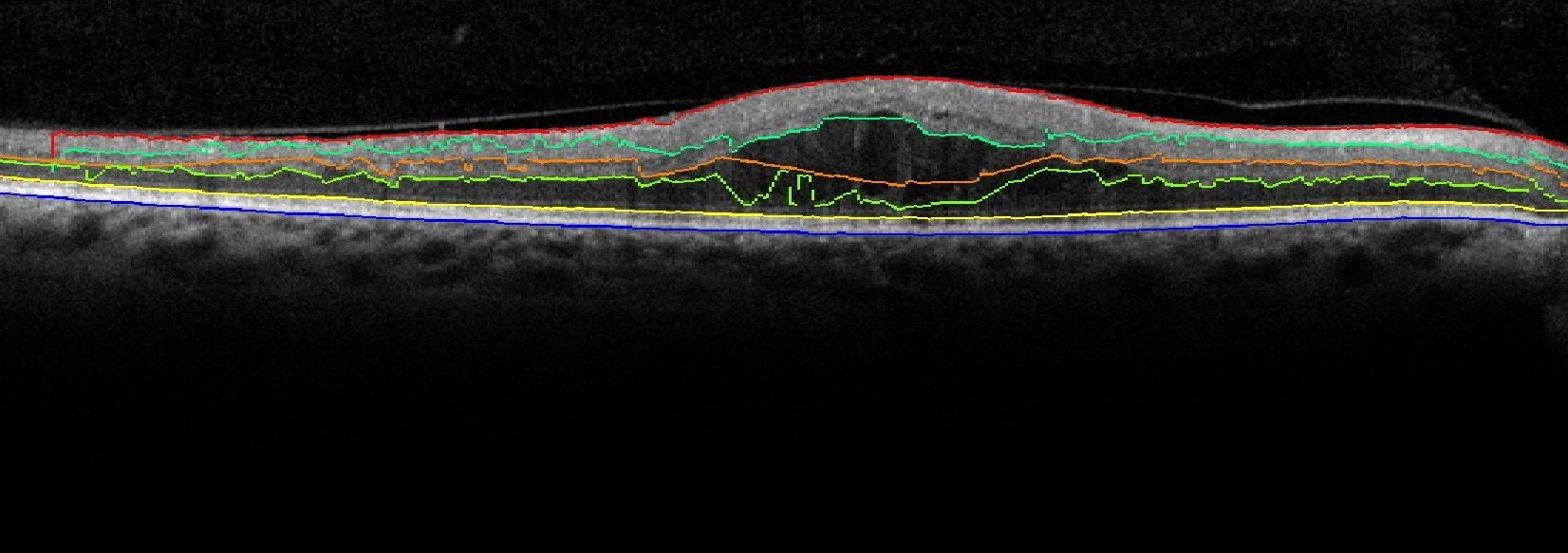}}
\subfigure[]{\includegraphics[width = 3.0in,height=1.0in]{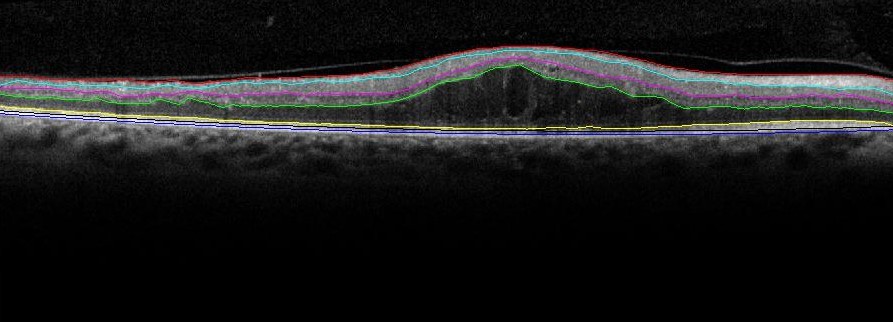}}
\subfigure[]{\includegraphics[width = 3.0in,height=1.0in]{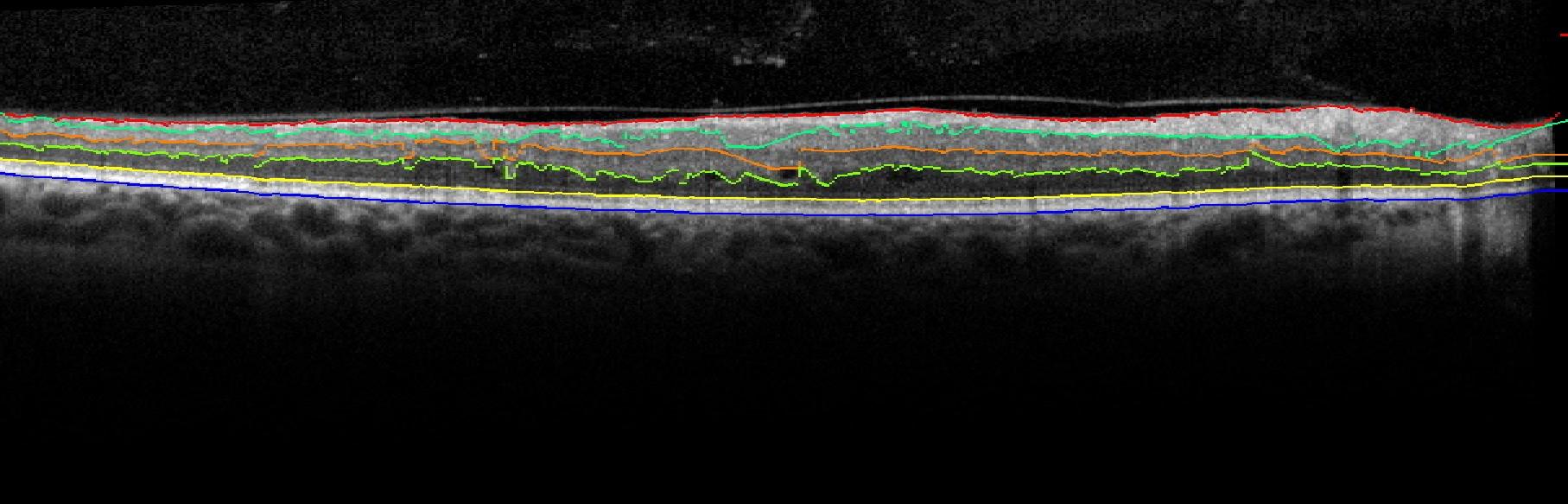}}
\subfigure[]{\includegraphics[width = 3.0in,height=1.0in]{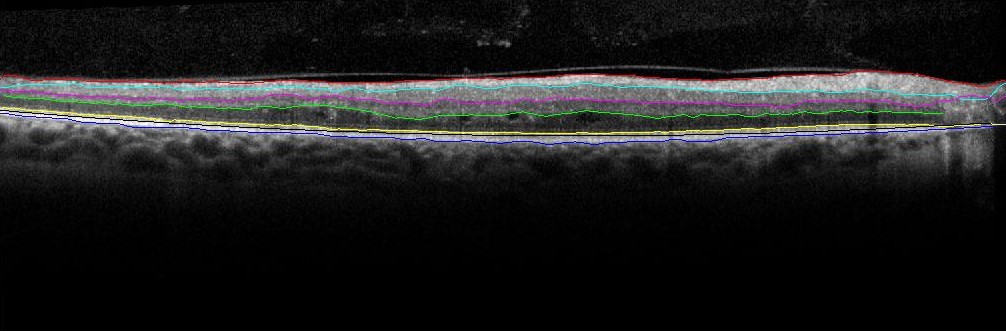}}
\subfigure[]{\includegraphics[width = 3.0in,height=1.0in]{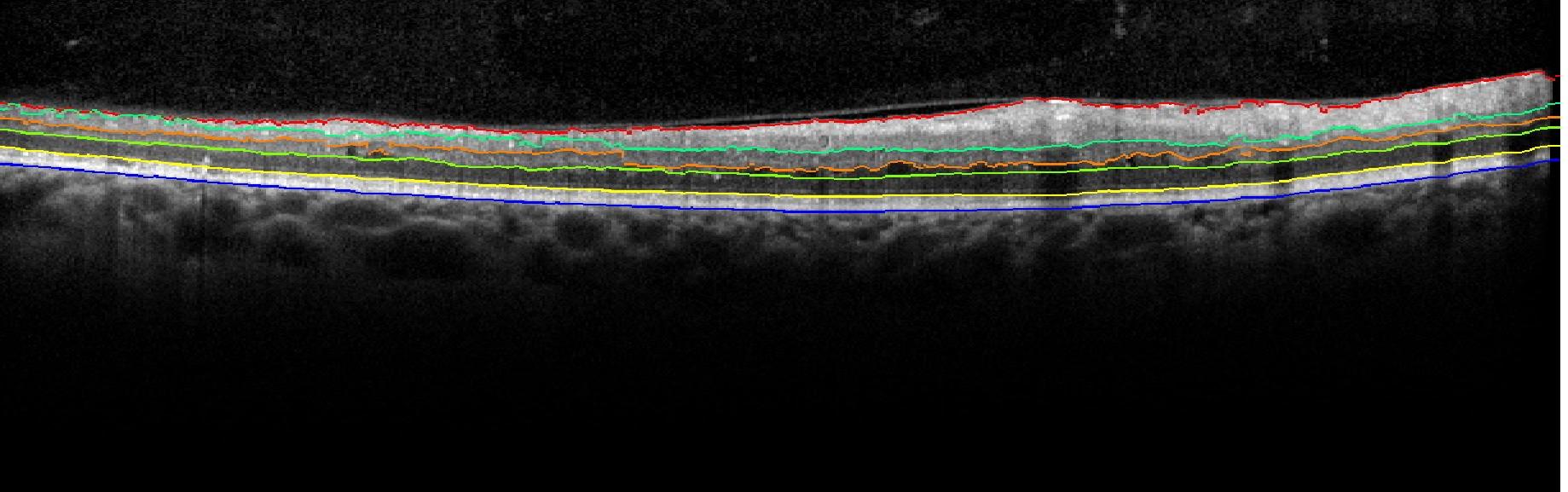}}
\subfigure[]{\includegraphics[width = 3.0in,height=1.0in]{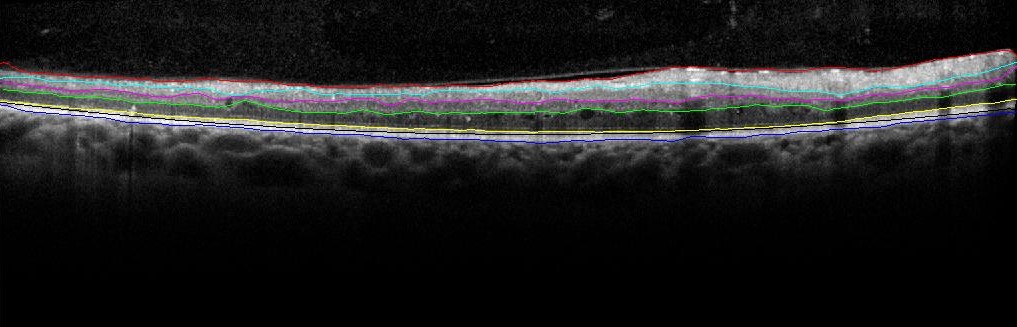}}
\caption{Comparative assessment of sub-retinal surface segmentation using the proposed method and the OCTSEG system on images with DME. Left column represents the automated segmentation produced by OCTSEG. Right column represents automated segmentation produced by the proposed method.}\label{seg_res}
\end{center}
\end{figure}

Since the error in Surface 2 is significantly high for abnormal  OCT images using the proposed segmentation algorithm, the sub-retinal layer thicknesses on such abnormal images will be analyzed for the NFL and IPL layers combined (NFL$+$IPL, between Surface 1 and Surface 3), the INL, ONL and Outer layer (between Surface 5 and Surface 7). It is noteworthy, that although the performance of automated surface segmentation deteriorates on images with pathology, the proposed denoising and segmentation method has more reliable and repeatable performance for inner sub-retinal surface segmentation when compared to the OCTSEG system as shown in Fig. \ref{seg_res}. The proposed method is capable of reliable estimation of the INL and ONL in image with cysts and sub-retinal disorganizations.

\subsection{Sub-retinal Layer Thickness Estimation}
Once the sub-retinal surfaces are segmented, the next step is to estimate average sub-retinal layer thicknesses. The mean and standard deviation in the sub-retinal layer thicknesses for normal and abnormal OCT images are shown in Table \ref{thickness_norm} and Table \ref{thickness_dme}, respectively. In these tables the mean thicknesses of every sub-retinal layer segmented manually are compared with the automated sub-retinal layer thickness estimation results using the proposed method and the CDWT approach. The sub-retinal layer segmentation distribution is analyzed in terms of the average layer thickness, mean correlation coefficient `r' and $R^2$ statistic between the manual and automated thickness measurements made for each image at every horizontal pixel location. A higher value of `r' and $R^2$ statistic for each OCT image denotes better sub-retinal layer segmentation performance.
\begin{table}[ht]
\begin{center}
\caption{Mean and standard deviation of sub-retinal layer thickness in normal images measured in $\mu m$.}
\begin{tabular}{|c |c |c |c|}
\hline 
Layer&Manual&Proposed(r,$R^2$)&CDWT (r,$R^2$)\\ \hline
NFL&52.4$\pm$8.7&52.6$\pm$11.5(0.86,0.7)&49.4$\pm$12.2(0.62,0.4)\\ 
IPL/GL&72.7$\pm$5.6&73.3$\pm$9.3(0.79,0.6)&61.9$\pm$15.6 (0.49,0.2)\\ 
INL&57.42$\pm$7.4&59.9$\pm$10.3(0.83,0.7)&72.2$\pm$4.9(0.39,0.1)\\ 
ONL&74.97$\pm$6.0&73.1$\pm$6.8(0.83,0.7)&70.5$\pm$11.0(0.39,0.1)\\ 
IS/OS&38.02$\pm$10.5&38.2$\pm$11.0(0.77,0.6)&20.3$\pm$5.9(0.36,0.1)\\ 
RPE&35.01$\pm$6.6&35.5$\pm$10.0(0.75,0.6)&42.0$\pm$8.3(0.33,0.1)\\ \hline
NFL$+$IPL&225.9$\pm$9.1&226.9$\pm$9.9(0.98,0.9)&226.4$\pm$6.39(0.43,0.2)\\
Inner&244.6$\pm$10.8&245.8$\pm$11.1(0.98,0.9)&250.3$\pm$8.9(0.32,0.1)\\
Outer&64.5$\pm$7.1&65.1$\pm$8.64(0.82,0.7)&53.70$\pm$6.3(0.39,0.1)\\ \hline
\end{tabular}
	\label{thickness_norm}
	\end{center}
\end{table}
In Table \ref{thickness_norm}, we observe that for OCT images from the normal set, the proposed method incurs up to 2 $\mu m$ error in estimating the mean sub-retinal layer thickness with $r, R^2$ consistently greater than 0.75 and 0.6, respectively. However, the CDWT method incurs $3-18 \mu m$ error in estimating the sub-retinal layer thickness with significantly low $r, R^2$ metrics. The p-values for the proposed denoising segmentation system and the CDWT denoising followed by the proposed segmentation system are $p<0.001$ and $p<0.01$, respectively. This analysis shows that the errors between the proposed denosinig approach and CDWT approach for mean sub-retinal layer thicknesses and distributions of sub-retinal layer thickness estimations are significantly higher than the errors in segmenting sub-retinal surfaces. 
\begin{table}[ht]
\begin{center}
\caption{Mean and standard deviation of sub-retinal layer thickness in abnormal images measured in $\mu m$.}
\begin{tabular}{|c |c |c |c|}
\hline 
Layer&Manual&Proposed(r,$R^2$)&CDWT(r,$R^2$)\\ \hline
NFL$+$IPL&111.5$\pm$15.9&129.4$\pm$19.7(0.78,0.6)&135.6$\pm$29.7(0.75,0.6)\\ 
INL&62.7$\pm$10.6&60.9$\pm$15.0(0.73,0.5)&76.9$\pm$14.5(0.36,0.1)\\ 
ONL&89.9$\pm$15.3&81.0$\pm$13.9(0.71,0.5)&70.9$\pm$11.4(0.23,0.1)\\ \hline
Inner&265.6$\pm$33.1&274.9$\pm$34.6(0.92,0.8)&274.0$\pm$61.3(0.90,0.8)\\
Outer&60.9$\pm$8.44&57.4$\pm$14.4(0.74,0.6)&49.7$\pm$10.6(0.32,0.1)\\ \hline
\end{tabular}
	\label{thickness_dme}
	\end{center}
\end{table}
In Table \ref{thickness_dme}, we observe that the proposed method incurs $2-18 \mu m$ error in estimating the mean sub-retinal layer thickness with $r, R^2$ statistics consistently greater than 0.7 and 0.5, respectively. On the other hand, the CDWT method incurs $12-24\mu m$ error for mean thickness estimation with $(r, R^2)$ statistics ranging from (0.3, 0.1) to (0.9, 0.8). 

\section{Conclusions and Discussion}\label{conclusion}
In this work we have presented a CAD system that denoises and segments 7 sub-retinal surfaces and 6 sub-retinal layers in the retinal micro-structure for normal OCT images from healthy patients and in abnormal images from patients with DME in less than 35 seconds per image. The proposed system is implemented on a 2.53 GHz Intel Core i3 and 3 GB RAM Laptop system using MATLAB. Also, the sub-retinal layer thickness distributions from all images belonging to an OCT image stack are combined to generate thickness maps that aid automated assessment of the progression in DME with time.

We analyze the performance of OCT image noise removal based on the proposed Wiener deconvolution method that estimates noise parameters using a Fourier-domain structural error metric. The proposed denoising method is compared to wavelet-transform based CDWT method by Chitchian et al. \cite{Erik}. Our analysis shows that the proposed Fourier domain-based denoising method improves the image SNR by more than 12dB, and it retains the sub-retinal surface edges within the retinal micro-structure while suppressing the noise significantly in the image background, thereby incurring a PSNR of about 23.5dB. The CDWT method on the other hand conserves the sub-retinal structure of the image foreground and background by smoothing the sub-retinal edges, thereby achieving a high PSNR of about 28dB but an SNR enhancement of less than 5dB. Further analysis shows that the CDWT denoising method incurs more error while automated segmentation of sub-retinal surfaces when compared to the proposed denoising method. Thus, the proposed Fourier-domain based denoising method is an important first step for the proposed sub-retinal surface and layer segmentation algorithm.

Another key observation is that the segmentation performance of the proposed denoising and segmentation algorithm deteriorates significantly from normal OCT images to abnormal ones, i.e., errors in mean sub-retinal thickness estimations are $0-2 \mu m$ for normal images and $2-18 \mu m$ for abnormal images. However, the CDWT denoising method followed by the proposed segmentation significantly deteriorate the estimations of the distributions of sub-retinal thicknesses (low `r' value), but the estimations of mean sub-retinal layer thicknesses do not deteriorate as much, i.e., errors in mean sub-retinal thickness estimations are $3-18\mu m$ for normal images and $12-24 \mu m$ for abnormal images. These observations demonstrate that the proposed segmentation algorithm is robust and can be combined with other denoising approaches to estimate mean sub-retinal layer thickness in normal and abnormal images. 

The proposed sub-retinal surface segmentation algorithm exhibits significant reliability (in terms of correlation coefficient (r) and $R^2$ metric) in extracting sub-retinal layer thicknesses for normal and abnormal OCT image stacks. Such a CAD system can be useful for detecting and monitoring the progression of pathology in DME patients, and in guiding clinical research directed towards treatment protocols. Future efforts will be directed towards developing CAD systems for automated segmentation of OCT image stacks from patients with other vision threatening pathologies that cause variations in sub-retinal layer thicknesses such as Glaucoma and Macular Telangiectasia (MacTel).

\bibliographystyle{IEEEtran}
\bibliography{IEEEabrv,references}

\end{doublespace}

\end{document}